\documentclass[letterpaper, 10 pt, conference]{ieeeconf}  

\IEEEoverridecommandlockouts                              

\usepackage[numbers]{natbib}
\usepackage{graphicx}
\usepackage{amsmath}
\usepackage{hyperref}
\usepackage{algpseudocode}
\usepackage{algorithm}
\usepackage[algo2e]{algorithm2e}
\usepackage{multirow}
\usepackage[normalem]{ulem}
\usepackage{booktabs}
\usepackage{gensymb}
\usepackage{textcomp}
\usepackage{lipsum}
\usepackage[font=small, hypcap=false]{caption}

\usepackage{amsmath,amsfonts,bm}
\usepackage{caption}
\usepackage{subcaption}

\def\eqref#1{equation~\ref{#1}}

\def\1{\bm{1}}

\DeclareMathAlphabet{\mathsfit}{\encodingdefault}{\sfdefault}{m}{sl}
\SetMathAlphabet{\mathsfit}{bold}{\encodingdefault}{\sfdefault}{bx}{n}

\newcommand{\xxnote}[3]{}
\ifx\hidenotes\undefined
  \renewcommand{\xxnote}[3]{\color{#2}{#1: #3}}
\fi

\hypersetup{
    colorlinks=true,
    linkcolor=blue,
    filecolor=magenta,      
    citecolor=blue,
    urlcolor=blue,
}

\renewcommand\thefigure{\arabic{figure}}
\newcommand{\method}{\textsc{HuDOR}}
\newcommand{\website}{\url{https://object-rewards.github.io/}}

\title{\LARGE \bf
Bridging the Human to Robot Dexterity Gap through \\Object-Oriented Rewards
}

\author{
Irmak Guzey$^{\dagger}$ \qquad Yinlong Dai \qquad Georgy Savva \qquad Raunaq Bhirangi \qquad Lerrel Pinto
\\ \\ New York University
\\ \\ { \tt \href{https://object-rewards.github.io/}{object-rewards.github.io}}\\
\thanks{$^{\dagger}$Correspondence to \texttt{irmakguzey@nyu.edu}.}
}

\newboolean{include-notes}
\setboolean{include-notes}{true}

\begin{document}

\makeatletter
\let\@oldmaketitle\@maketitle%
\renewcommand{\@maketitle}{\@oldmaketitle%
    \centering
    \includegraphics[width=\linewidth]{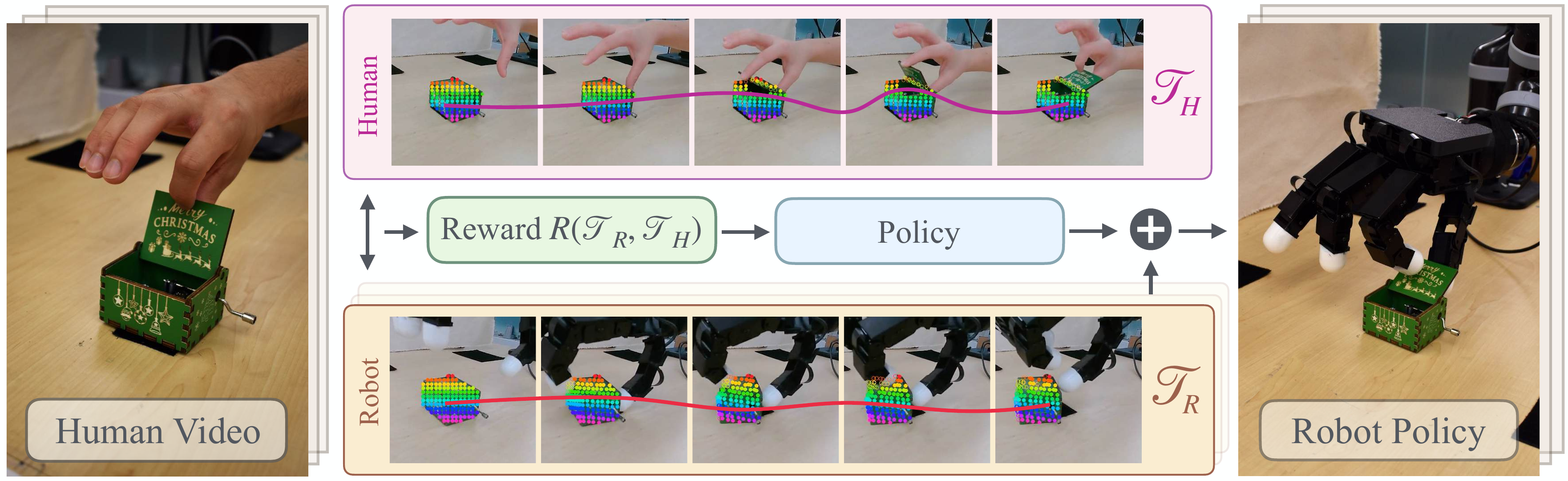}
    \captionof{figure}{\method{} generates rewards from human videos by tracking points on the manipulable object, indicated by the rainbow-colored dots, over the trajectory. This allows for online training of multi-fingered robot hands given only a \textit{single} video of a human solving the task (left) without any robot teleoperation. To optimize the robot's policy (middle), rewards are computed by matching the point movements of the robot policy $\mathcal{T}_R$ with those in the human video $\mathcal{T}_H$. In under an hour of online fine-tuning, our Allegro robot hand (right) is able to \textit{open the music box}.}
    \label{fig:intro}
}

\makeatother

\maketitle
\thispagestyle{empty}
\pagestyle{empty}

\begin{abstract}

Training robots directly from human videos is an emerging area in robotics and computer vision. While there has been notable progress with two-fingered grippers, learning autonomous tasks for multi-fingered robot hands in this way remains challenging. A key reason for this difficulty is that a policy trained on human hands may not directly transfer to a robot hand due to morphology differences. In this work, we present \method{}, a technique that enables online fine-tuning of policies by directly computing rewards from human videos. Importantly, this reward function is built using object-oriented trajectories derived from off-the-shelf point trackers, providing meaningful learning signals despite the morphology gap and visual differences between human and robot hands. Given a single video of a human solving a task, such as gently opening a music box, \method{} enables our four-fingered Allegro hand to learn the task with just an hour of online interaction. Our experiments across four tasks show that \method{} achieves a 4$\times$ improvement over baselines. Code and videos are available on our website, \website{}.

\end{abstract}

\setcounter{figure}{1}


\section{Introduction}
\label{sec:intro}

Humans effortlessly perform a wide range of dexterous tasks in their daily lives~\cite{kumar2016dex}. Achieving similar capabilities in robots is essential for their effective deployment in the real world. 
Towards this end, recent advances have enabled the learning of multi-modal, long-horizon, and dexterous behaviors for two-fingered grippers \cite{zhao2023act, lee2024vqbet, chi2024diffusionpolicy, cui2022play} using imitation learning (IL) from teleoperated robot data. However, extending such methods to complex tasks with multi-fingered hands has proven challenging. 

The challenge of using teleoperation-based learning for multi-fingered hands arises from two key issues. First, achieving even moderate robustness this way requires large amounts of data. Tasks involving two-fingered grippers \cite{brohan2023rt, open2024x, etukuru2024rum, haldar2024baku} often demand thousands of demonstrations to train robust policies. This data requirement is likely even greater for hands with larger action dimensions and tasks that require higher levels of precision and dexterity. Second, teleoperating multi-fingered hands presents a challenging systems problem due to the need for low-latency and continuous feedback when controlling multiple degrees of freedom~\cite{arunachalam2022holodex, iyer2024openteach, ding2024pro}. This makes it even harder to collect the large amounts of data needed to accomplish dexterous tasks.


An alternate approach that circumvents teleoperation is to develop policies for robots using videos of humans executing tasks~\cite{kumar2022graph, smith2019avid, eze2024lbw}. However, most previous approaches have required either additional teleoperated robot demonstrations ~\cite{wang2023mimic} or human-intervened learning~\cite{wang2024dexcap} for fine-tuning. This extra information is often necessary to bridge the gap between the morphological and visual differences between human hands (as seen in human video data) and robot hands (as observed in robot interactions).



In this work, we present \method{}, a new approach to bridge the gap between human videos and robot policies through online imitation learning. Given a human video and hand pose trajectory, an initial robot replay can be generated using pose transformation and full robot inverse kinematics. However, this initial replay often fails due to morphological differences between human and robot hands. \method{} improves this initial replay through a multi-step process: (a) We track the points of the manipulated object in both human and robot trajectory videos; (b) then compute the similarity of object motion and articulation using these tracked point sets, (c) and finally fine-tune the initial robot policy through reinforcement learning. By iteratively refining its performance based on these comparisons, the robot effectively imitates human demonstrations while adapting to its physical constraints. Our framework is shown in Fig. \ref{fig:intro}.

We evaluate \method{} on four dexterous tasks, including opening a small music box with one hand and slide-picking up a thin card. Our contributions can be summarized as follows:
\begin{enumerate} 
    \item \method{} introduces the first framework that enables the learning of dexterous policies on multi-fingered robot hands using only a single human video and hand pose trajectory (Section \ref{sec:experiments}).
    \item \method{} introduces a novel approach for object-oriented reward computation that matches human and robot trajectories. This results in 2.1$\times$ better performance on three of our tasks than common reward functions (Section~\ref{sec:experiments:reward}).
    \item \method{} outperforms state-of-the-art offline imitation learning methods for learning from human demonstrations \cite{wang2024dexcap, lee2024vqbet}, achieving an average improvement of 2.64$\times$, emphasizing the need for online corrections (Section~\ref{sec:experiments:online}). 
\end{enumerate}

Robot videos are best viewed on our website: \website{}.
\section{Related Works}
\label{sec:related}
Our work draws inspiration from extensive research in dexterous manipulation, learning from human videos, and imitation learning. In this section, we focus our discussion on the most pertinent contributions across these interrelated areas.

\paragraph{Robot Learning for Dexterous Manipulation}

Learning dexterous policies for multi-fingered hands has been a long-standing challenge that has captured the interest of the robotics learning community \cite{handa2022dextreme, openai2019rubik, arunachalam2022holodex}. Some works have addressed this problem by training policies in simulation and deploying them in the real world \cite{shaw2023leap, openai2019rubik}. Although this approach has produced impressive results for in-hand manipulation~\cite{yin2023rotation, ma2023eureka}, closing the sim-to-real gap becomes cumbersome when manipulating in-scene objects.

Other works have focused on developing different teleoperation frameworks~\cite{arunachalam2022holo, yang2024ace, iyer2024openteach, DexPilot} and training offline policies using robot data collected through these frameworks. While these frameworks are quite responsive, teleoperating dexterous hands without directly interacting with objects remains difficult for users due to the morphological mismatch between current robotic hands and the lack of tactile feedback for the teleoperator.

Given the challenges of large-scale data collection, most offline dexterous policies tend to fail due to overfitting. To mitigate this, some previous works have focused on learning policies with limited data~\cite{guzey2023dexterity, haldar2022watch}, either by using nearest-neighbor matching for action retrieval~\cite{pari2021surprising, guzey2023dexterity, iyer2024openteach, arunachalam2022holo} or by initializing with a sub-optimal offline policy and finetuning that or learning a residual policy with online interactions to improve generalization~\cite{haldar2022watch, guzey2023tavi, haldar2023fish}.

\paragraph{Learning from Human Videos}
To scale up data collection using more accessible sources, the vision and robotics communities have worked on learning meaningful behaviors and patterns from human videos \cite{liang2024dream, yang2023learning, karl2022skill, grauman2023ego}. Some efforts focus on learning simulators that closely mimic the real-world environment of the robot from human videos using generative models~\cite{yang2023learning, julen2024generative, liang2024dream}, using these simulators to train policies and make decisions by predicting potential future outcomes.

Other works use internet-scale human videos to learn higher-level skills or affordances~\cite{karl2022skill, bahl2023affordance}.  However, these works either require low-level policies to learn action primitives for interacting with objects~\cite{karl2022skill} or only focus on simple tasks where a single point of contact is sufficient for manipulation~\cite{bahl2023affordance}. Yet other approaches leverage on-scene human videos to learn multi-stage planning~\cite{wang2023mimic, smith2019avid} but need additional robot data to learn lower-level object interactions. Notably, all of these works focused on two-gripper robots, where manipulation capabilities are limited and objects are less articulated.

A few recent studies \cite{wang2024dexcap, chen2024arcap, li2024okami} address this issue for dexterous hands by using in-scene human videos collected by multiple cameras in conjunction with hand motion capture systems. These studies either focus on simple grasping tasks~\cite{li2024okami, chen2024arcap} or require an online fine-tuning stage with human feedback for dexterous tasks~\cite{wang2024dexcap}. Additionally, the offline learning process for these approaches requires extensive pre-processing to mask the human hand from the environment point cloud. 

\method{} differs from these works by eliminating the need for cumbersome human feedback, automatically extracting a reward from a single human demonstration, and allowing the robot to learn from its mistakes to account for the morphology mismatch between the robot and the human.


\section{Learning Teleoperation-Free Online Dexterious Policies }

\label{sec:method}

\method{} introduces a framework to learn dexterous policies from a single in-scene human video of task execution. Our method involves three steps: (1) A human video and corresponding hand pose trajectory are recorded using a VR headset and an RGB camera; (2) hand poses are transferred and executed on the robot using pose transformation and full-robot inverse kinematics (IK); and (3) reinforcement learning (RL) is used to successfully imitate the expert trajectory. In this section, we explain each component in detail.

\subsection{Robot Setup and Human Data Collection}
\label{sec:method:setup}

\begin{figure}[t]
    \centering
    \includegraphics[width=\linewidth]{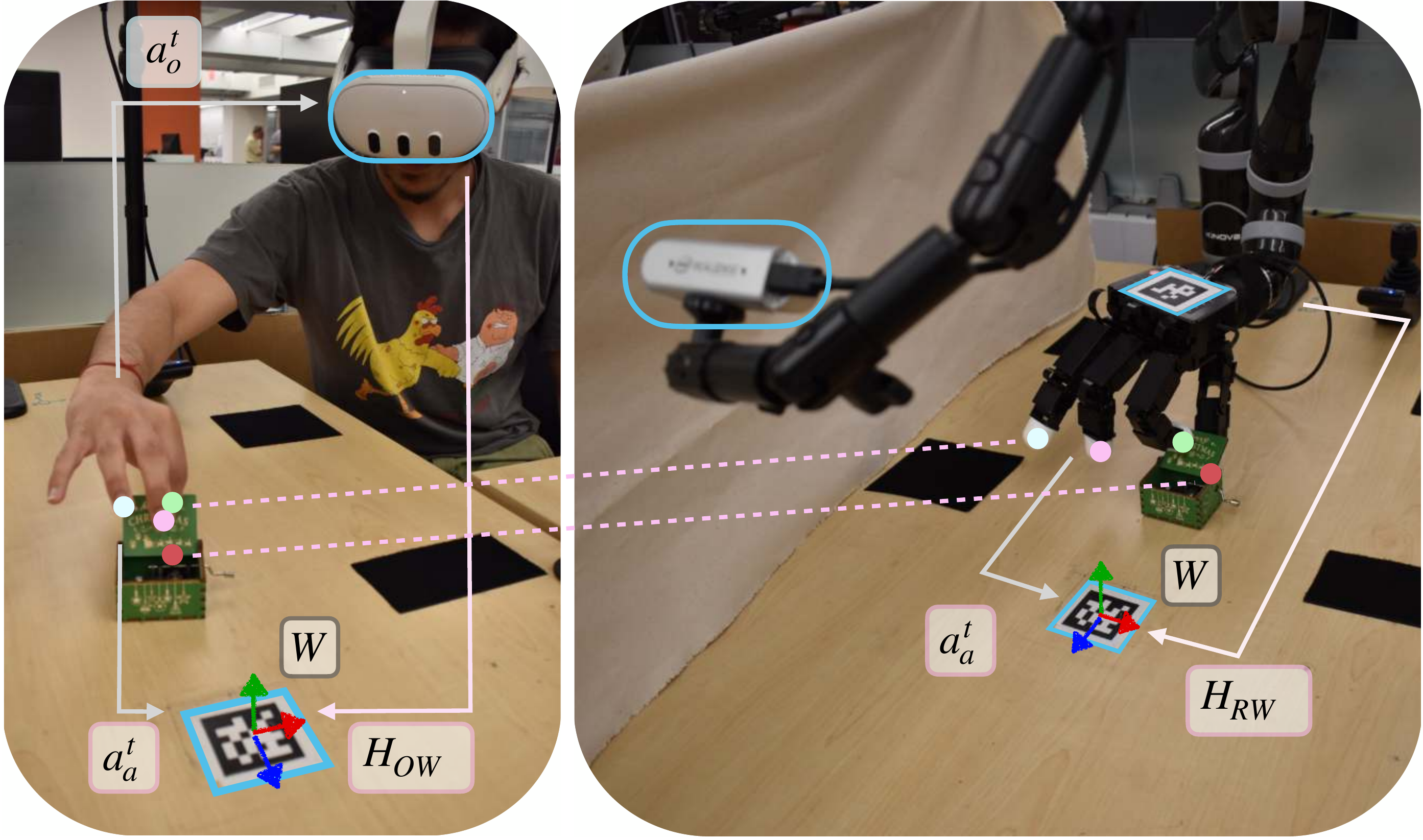}
    \caption{Illustration of the robot setup and trajectory transfer in \method{}. ArUco markers are used for calibration. The human demonstration is collected in-scene, i.e. demonstrator is in the same scene as the robot. The VR headset is used solely for obtaining the fingertip positions with respect to the robot frame (illustrated with colored dots) and can be worn or attached to the setup as needed.  World frame $W$ is visualized on the ArUco marker on the operation table.}
    \label{fig:robot_setup}
    \vspace{-15pt}
\end{figure}

Our hardware setup includes a Kinova JACO 6-DOF robotic arm with a 16-DOF four-fingered Allegro hand \cite{arunachalam2022holodex} attached. Two RealSense RGBD cameras \cite{leonid2017realsense} are positioned around the operation table for calibration and visual data collection. A Meta Quest 3 VR headset is used to collect hand pose estimates. Our first step involves computing the relative transformation between the VR frame and the robot frame to directly transfer the recorded hand pose trajectory from the human video to the robot as shown in Fig.~\ref{fig:robot_setup}. We use two ArUco markers -- one on the operation table and another on top of the Allegro hand -- to compute relative transformations. The first marker is used to define a world frame and transform fingertip positions from the VR frame to the world frame, while the second marker is used to determine the transformation between the robot's base and the world frame. 

\paragraph{Relative Transformations} We collect human hand pose estimates using existing hand pose detectors on the Quest 3 VR Headset, and capture visual data using the RGBD cameras. Fingertip pose for the $i^{\text{th}}$ human fingertip captured in the VR frame at time \( t \), \( a^{t,i}_o \), are first transformed to the world frame as \( a^{t,i}_w = H_{OW} ~ a^{t,i}_o \), where \( H_{OW} \) is the homogeneous transform from the VR frame to the world frame. This transform is computed by detecting the ArUco marker on the table using the cameras on the VR headset. A standard calibration procedure \cite{hartley2002multiple} is used to compute the transformation \( H_{RW} \) between the robot frame and the world frame by detecting the two ArUco markers using the RGB camera shown in Fig.~\ref{fig:robot_setup}. This calibration allows us to directly transfer human fingertip positions from the Oculus headset to the robot's base using the equation:

\begin{align}
    a^{t,i}_r & = H_{RW}^{-1} ~ a^{t,i}_w \\
                & = H_{RW}^{-1} ~ H_{OW} ~ a^{t,i}_o
\end{align}

\noindent where, \( a^{t,i}_r \) are the homogeneous coordinates of the $i^{\text{th}}$ human fingertip positions from the recorded video in the robot frame. Henceforth, we use \( a^t_r = [a^{t,0}_r, a^{t,1}_r, a^{t,2}_r a^{t,3}_r] \) to refer to the 12-dimensional vector containing concatenated locations of the four fingertips in robot frame.

\paragraph{Data Collection} Using the VR application we implemented, a user can pinch the index finger and thumb of their right hand to begin interacting with the object directly using their own hands. After collecting the demonstration, they can pinch their fingers again to signal the end of the demo. We calculate the wrist pose relative to the world frame at the start of the demo, and during deployment, we initialize the robot's arm to that initial wrist pose. This allows the robot to begin its exploration from a suitable starting position. 

\paragraph{Data alignment} During data collection, we record the fingertip positions \( a^t_r \) and image data \( o^t \) for all $t = 1 \dots T$ where $T$ is the trajectory length. Since these components are collected at different frequencies, we align them on collected timestamps to produce synchronized tuples \( ( a^t_r, o^t) \) for each time \( t \). The data is then subsampled to 5 Hz.

\begin{figure*}[t]
    \centering
    \includegraphics[width=\linewidth]{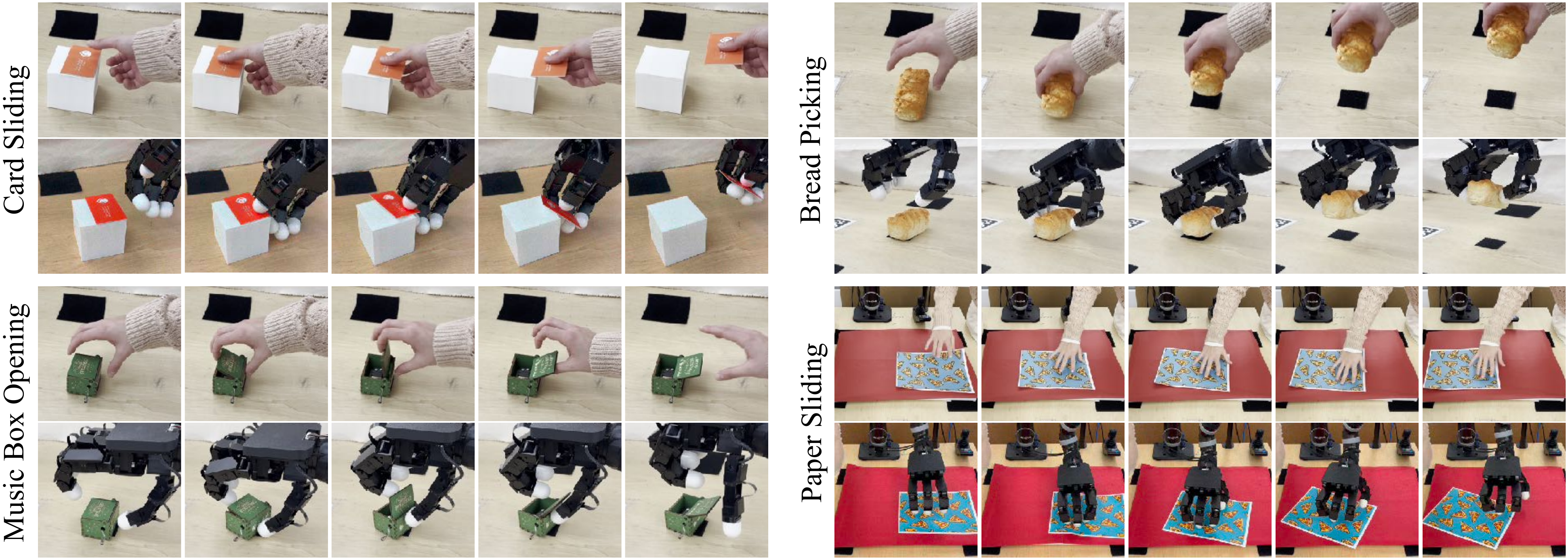}
    \caption{An illustration of the human demonstrations (top rows) and the corresponding robot policies (bottom rows) trained using \method{}. Our method does not require teleoperated robot data and learns to imitate human demonstrations through online interactions. Note the differences in hand motions between the learned robot policy and the human videos, reflecting the morphological differences.}
    \label{fig:example_human_figre}
    \vspace{-2mm}
\end{figure*}

\begin{figure}[t]
    \centering
    \includegraphics[width=\linewidth]{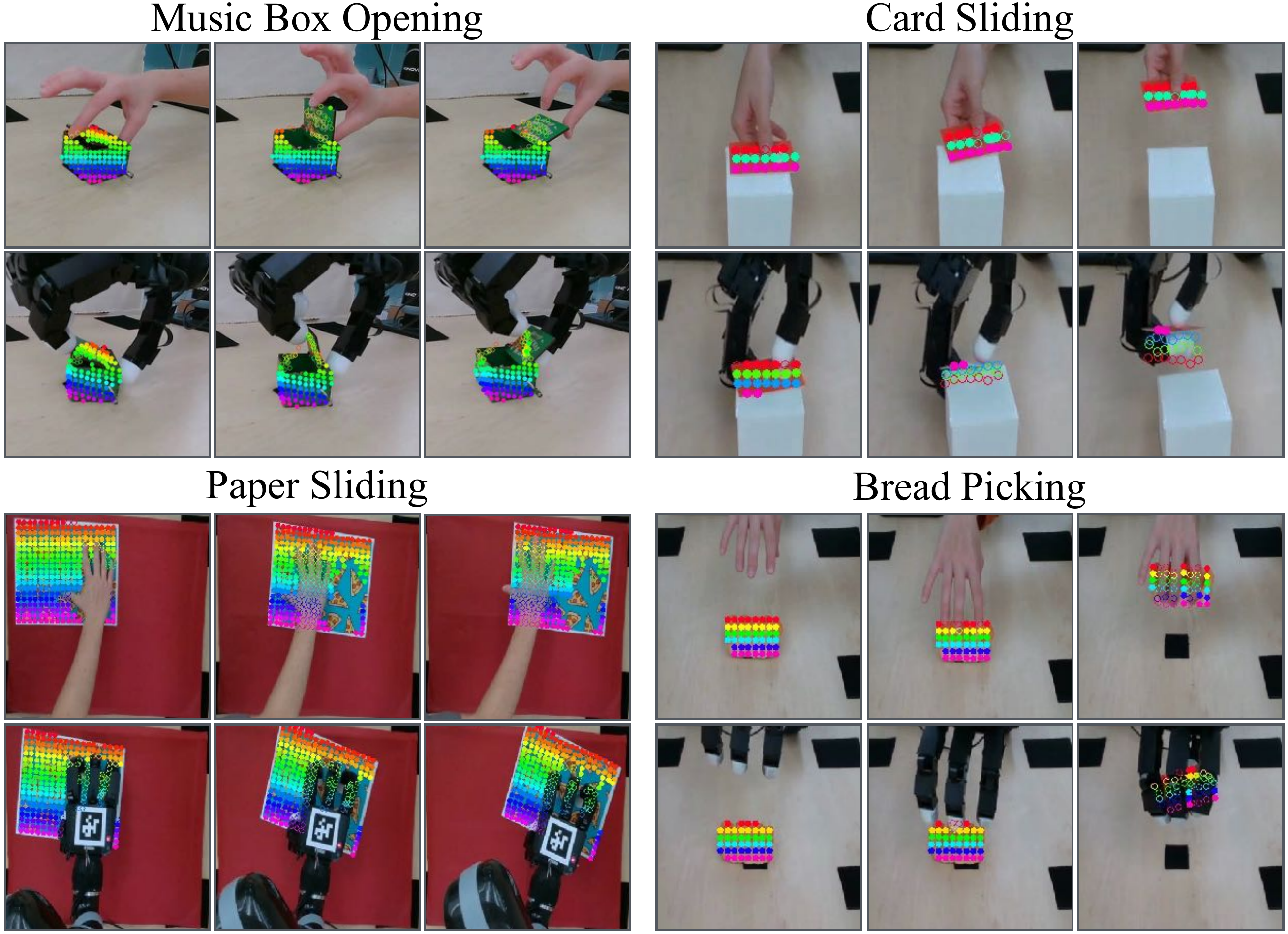}
    \caption{An illustration showing how masked objects appear in both the robot and human videos. Points on the objects represent tracking in each video. Occlusions are indicated by hollow points rather than solid ones.}
    \label{fig:masks_figure}
    \vspace{-4mm}
\end{figure}

\paragraph{Inverse Kinematics} To ensure the robot’s fingertips follow the desired positions relative to its base, we implement a custom inverse kinematics (IK) module for the full robot arm-hand system. This module uses gradient descent on the joint angles, using the Jacobian of the robot fingertips with respect to the joint angles, to minimize the distance between desired fingertip positions and the current ones~\cite{tarun2019convex}. For the IK optimizer, we apply different learning rates for the hand and arm joints, allowing it to prioritize the hand movements. The hand learning rate is set to be 50 times higher than the arm learning rate, enabling more natural and precise control of the fingers. To summarize, the IK module takes the desired fingertip positions \( a^t_r \) and the current joint positions of both hand and arm \( j^t \) as inputs, and outputs the next joint positions \( j^{\ast t+1} = I( a^t_r, j^t) \) needed to reach the target.

Using the calibration and IK procedures outlined above, our robot arm-hand system can follow a fingertip trajectory directly from an in-scene human video. We showcase the human demonstrations used in Fig. \ref{fig:example_human_figre}.

\subsection{Residual Policy Learning}
\label{sec:method:policy_learning}

Due to the morphological differences between the human and robot, as well as errors in VR hand pose estimation, naively replaying the retargeted fingertip trajectories on the robot mostly does not successfully solve the task, even when the object is in the same location. To alleviate this problem, we learn an online residual policy using reinforcement learning (RL) to augment the trajectory replay. Traditional RL algorithms for real-world robots rely on reward functions derived from straightforward methods such as image-based matching rewards~\cite{haldar2023watch, haldar2023fish, guzey2023tavi} using in-domain demonstration data. However, due to the significant difference in the visual appearance of human and robot hands, these methods do not provide effective reward signals. To get around this domain gap, we propose a novel algorithm for object-centric trajectory-matching rewards.

\begin{figure*}[!t]
    \centering
    \includegraphics[width=\linewidth]{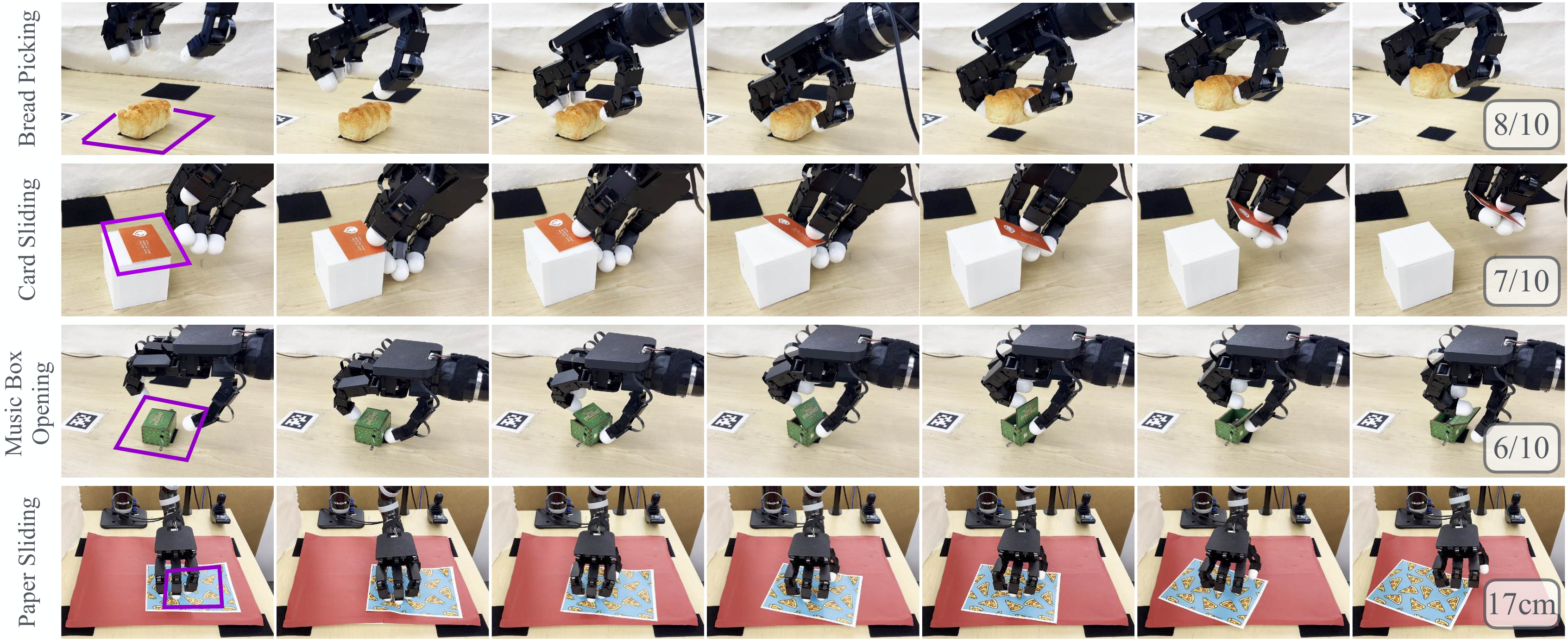}
    \caption{Rollouts of trained policies from \method{} on four tasks are shown. For all tasks, validation is performed at various locations within the illustrated areas in the leftmost frames, while training is conducted using a single human video where the initial object configuration is in the middle of these areas. Success for each task is shown in the rightmost frames. Videos are best viewed on our website: \website{}.}
    \label{fig:rollouts}
    \vspace{-15pt}
\end{figure*}

\paragraph{Object Point Tracking and Trajectory Matching}
\label{sec:method:reward_calculation}

Our reward computation involves using off-the-shelf computer vision models to track the motion of points on the object of interest. We compute the mean squared error between the 2D trajectories of these points in the human expert video and the robot policy rollout and use this as a reward at every timestep in our online learning framework. In this section, we explain our reward calculation in detail. 

\begin{itemize} 
    \item \textit{Object State Extraction}: Given a trajectory \(\tau = [o^1, \dots, o^T]\), where \(T\) is the length of the trajectory and \(o^t\) is an RGB image at time \(t\), we use the first frame \(o^1\) as input to a language-grounded Segment-Anything Model~\cite{medeiros2023langsegmentanything, kirillov2023sam} -- \texttt{langSAM}. \texttt{langSAM} uses a text prompt and GroundingDINO \cite{caron2021dino} to extract bounding boxes for the object, which are then input to the SAM \cite{kirillov2023sam} to generate a mask. The output of \texttt{langSAM} corresponding to $o^1$ is a segmentation mask for the initial object position, $P^1 \in \mathbb{R}^{N \times 2}$, which is represented as a set of $N$ detected points on the object, where $N$ is a hyperparameter. The parameter $N$ determines the density of object tracking and is adjusted based on the object's size in the camera view.

    \item \textit{Point Tracking}: The mask \(P^1\) is used to initialize the transformer-based point tracking algorithm Co-Tracker \cite{karaev2023cotracker}. Given a trajectory of RGB images, \(\tau\), and the first-frame segmentation mask, \(P^1\), Co-Tracker tracks points \(p_i^t = (x_i^t, y_i^t)\) in the image throughout the trajectory $\tau$ for all $t \in \{1 \dots T\}$, where $P^1 = [p_1^1, \dots p_N^1]$. We use $\tau^p = [P^1, \dots P^T]$ to denote the point trajectory consisting of the sets of tracked points. We illustrate what the tracking of the objects looks like for both human and robot trajectories in Fig. \ref{fig:masks_figure}. 

    \item \textit{Matching the Trajectories}: First, we define two additional quantities: centroid of the detected points, $\hat{P}^t$ and mean translation, $\delta^t_{trans}$ at time $t$. $\delta^t_{trans}$ is defined as the mean displacement of all points in $P^t$ from $P^1$. Concretely,

    \vspace{-4mm}
     \begin{align}
        \hat{P}^t = \frac{1}{N} \sum_{i=1}^N p^t_i \\
        \delta_{trans}^t = \hat{P}^t - \hat{P}^1 
    \end{align}

    We define the \textit{object motion} at time $t$ as $\mathcal{T}^t = \delta^t_{trans}$. Given two separate object motions at time $t$, one corresponding to the robot \(\mathcal{T}_R^t\) and one corresponding to the human \(\mathcal{T}_H^t\), the reward is calculated by computing the negative root mean squared error between them:
    
    \begin{equation}
        r^{H2R}_t = -\sqrt{\big(\mathcal{T}_R^t - \mathcal{T}_H^t\big) ^2} 
        \label{eq:reward}
    \end{equation}

\end{itemize}

\paragraph{Exploration Strategy}
\label{sec:method:exploration}
We select a subset of action dimensions to explore and learn from. For example, we focus only on the X and Y axes of the thumb for the Card Sliding task, rather than exploring all axes of all fingers. This approach speeds up the learning process and enables quick adaptation. Exploration axes for all tasks are mentioned in Section \ref{sec:experiments:tasks}. For the exploration strategy, we use a scheduled additive Ornstein-Uhlenbeck (OU) noise \cite{PhysRev.36.823, lillicrap2015continuous} to ensure smooth robot actions.

After extracting a meaningful reward function and identifying a relevant subset of the action space, we learn a residual policy $\pi_{r}(\cdot)$ on this subset by maximizing the reward function in Eq. \ref{eq:reward} for each episode using DrQv2~\cite{yarats2021mastering}.

Inputs to the residual policy $a^t_{+} = \pi_r()$ at time $t$ are (a) the human retargeted fingertip positions with respect to the robot's base $a^t_r$, (b) change in current robot fingertip positions $\Delta s^t = s^{t} - s^{t-1}$, (c) the centroid of the tracked points set on the robot trajectory $\hat{P}^t_R$ and (d) the object motion at time $t$, $\mathcal{T}^t_R$. Finally, we compute the executed actions as follows:

    \vspace{-4mm}
    \begin{align}
        a^t & = a^t_r + a^t_{+} \\ 
        & = a^t_r + \pi_r(a^t_r, \Delta s^t, \hat{P}^t_R, \mathcal{T}^t_R)
    \end{align}

The action, $a^t$, is sent to the IK module which converts it into joint commands for the robot. The policy is improved over time using DrQv2 as the robot accumulates experience interacting with the object.

\section{Experimental Evaluation}
\label{sec:experiments}

\begin{figure}[t]
    \centering
    \includegraphics[width=\linewidth]{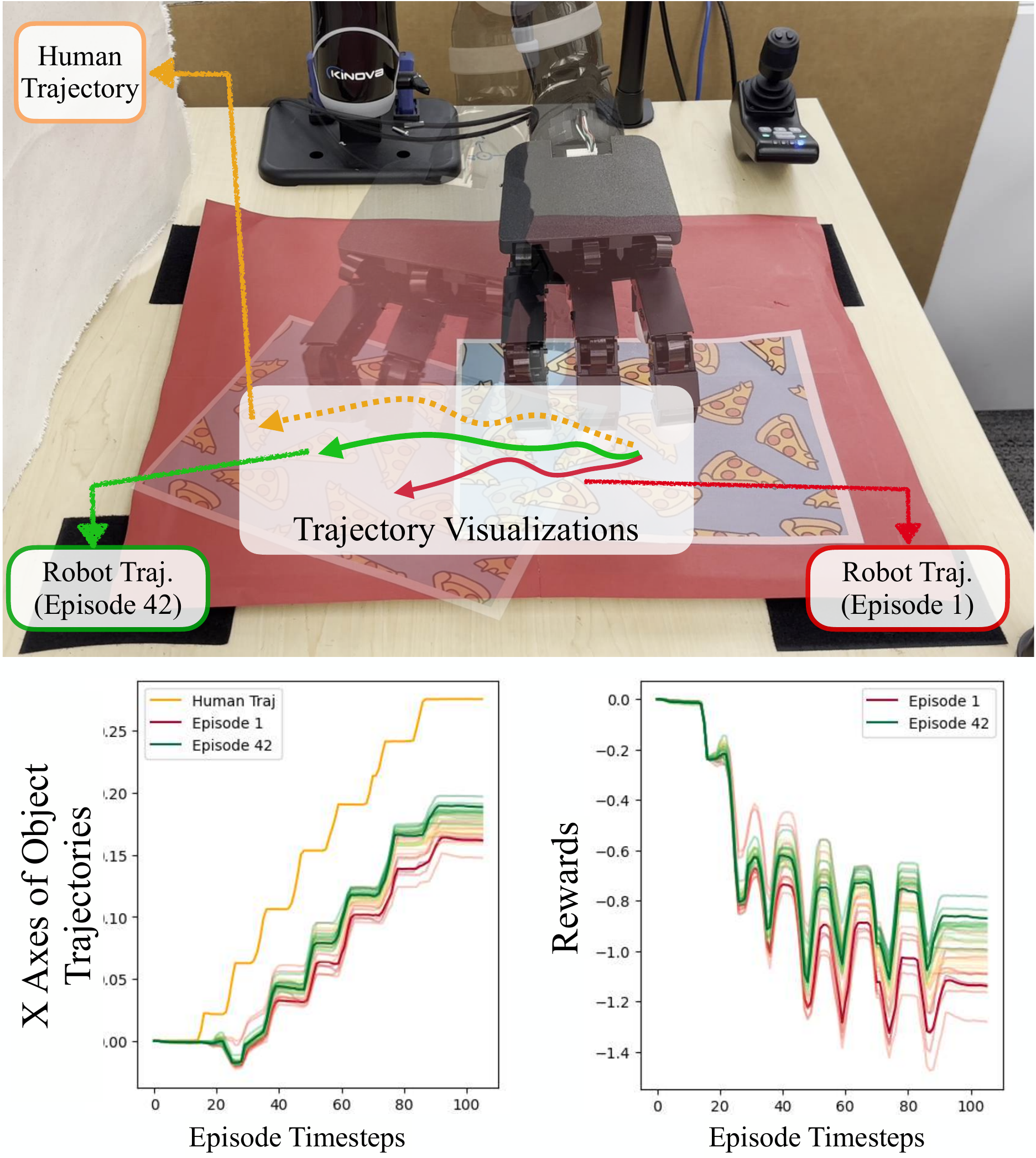}
    \caption{Illustration of how online correction improves robot policy and moves the robot trajectory closer to the expert's as time progresses in the Paper Sliding task. At the top, we visualize the trajectories of the paper in different episodes and the human video. At the bottom, we showcase the X-axis of the trajectory of the tracked points on the paper for different episodes and their corresponding rewards. The color of the episodes gradually changes from red in Episode 1 to green in Episode 42.}
    \label{fig:online_correction}
    \vspace{-15pt }
\end{figure}

\begin{figure}[t]
    \centering
    \includegraphics[width=\linewidth]{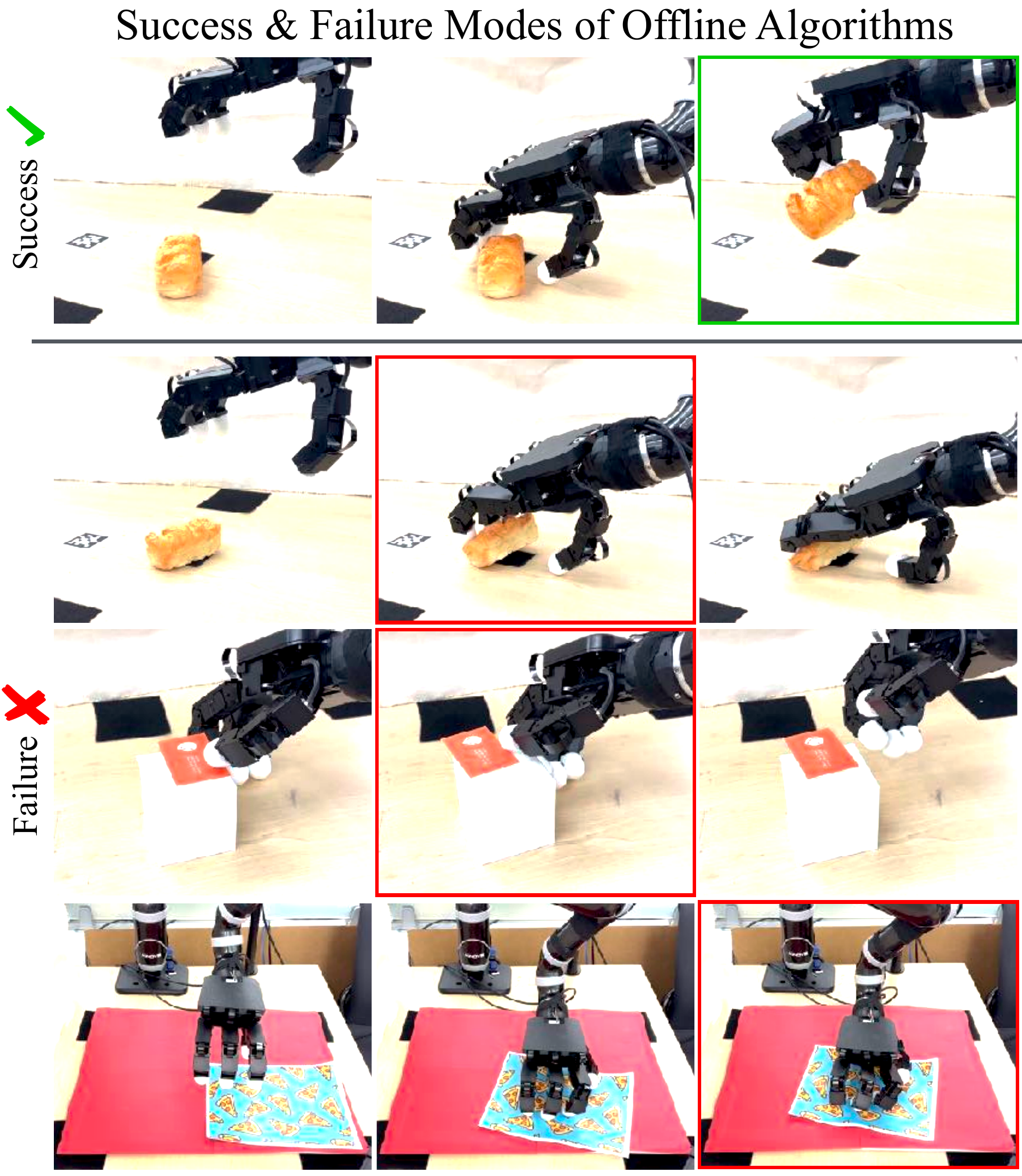}
    \caption{Illustration of failure modes in the Point Cloud BC algorithm. Red borders indicate frames, where the data goes out of distribution (OOD) and the algorithm, fails to recover. Note how, in the Bread Picking and Paper Sliding tasks, the hand progressively lowers, while in the Card Sliding task, once the card is slid to the edge, the algorithm fails to recover. In contrast, \method{} combines an open-loop base policy with a learned residual policy, providing more robust behavior against OOD cases.}
    \label{fig:failure_mode}
    \vspace{-15pt }
\end{figure}

We evaluate our method against 6 different baselines and run multiple ablations to answer the following questions: 
\begin{enumerate}
    \item How much do online corrections improve the performance of \method{}? 
    \item Does object-centric reward function in \method{} improve over common reward functions?
    \item How well does \method{} generalize to new objects and larger locations?
\end{enumerate}

\subsection{Task Descriptions} 
\label{sec:experiments:tasks}
We experiment with four dexterous tasks, which are visualized in Fig. \ref{fig:rollouts}. Exploration axes mentioned are with respect to the base of the robot. 

\paragraph{\textbf{Bread Picking}} The robot must locate an orange-colored piece of bread, pick it up, and hold it steadily for a sustained period. During validation, the bread is positioned and oriented within a 15cm $\times$ 10cm space. We explore the X axes of all fingers for this task. The text prompt used to retrieve the mask is \textit{orange bread}.

\paragraph{\textbf{Card Sliding}} The robot must locate and slide a thin card with its thumb and pick it up by supporting it with the rest of its fingers. During validation, the card is positioned and oriented within a 10cm $\times$ 10cm space. The text prompt used to retrieve the object mask is \textit{orange card}. We explore only the X and Y axes of the thumb.

\paragraph{\textbf{Music Box Opening}} The robot must locate and open a small music box. It uses its thumb to stabilize the box while unlatching the top with its index finger. During validation, the box is positioned and oriented within a 10cm × 10cm space. We explore all axes of the thumb and index fingers, and the text prompt used is \textit{green music box}. 

For this task, since the rotation of the object was significant, for each time $t$, we calculated the mean rotation vector about the centroid of all the points in $P^t$ from $P^1$, $\delta_{rot}^t = \frac{1}{N}\sum_{i=1}^N \left[ \left(p^t_i -  \hat{P}^t \right) \times \left(p^1_i - \hat{P}^1 \right) \right]$, in addition to the mean translation, $\delta^t_{trans}$. The final object motion is then calculated as $\mathcal{T}^t = [\delta^t_{trans}, \delta^t_{rot}]$ and used in Eq. \ref{eq:reward}. Additionally, we observed that a sparse reward was better for this task, so we only used the last 5 frames of the trajectory for reward calculation for \method{} and all of our baselines.

\paragraph{\textbf{Paper Sliding}} The robot must slide a given piece of paper to the right. During validation, the paper is positioned and oriented within a 15cm $\times$ 15cm space. The text prompt used to retrieve the mask is \textit{blue paper with pizza patterns}. Higher rewards are given as the paper moves further to the right. Success in this task is measured by the distance the paper moves to the right, expressed in centimeters. We explore on X and Z axes of all the fingers. 

\textit{Evaluating robot performance}: To compare the robot's performance, we evaluate \method{} against various online and offline algorithms. For all online algorithms, we train the policies until the reward converges, up to one hour of online interactions. We evaluate the methods by running rollouts on 10 varying initial object configurations for every task.

\subsection{How important are online corrections?}
\label{sec:experiments:online}

\begin{figure*}[t]
    \centering
    \includegraphics[width=\textwidth]{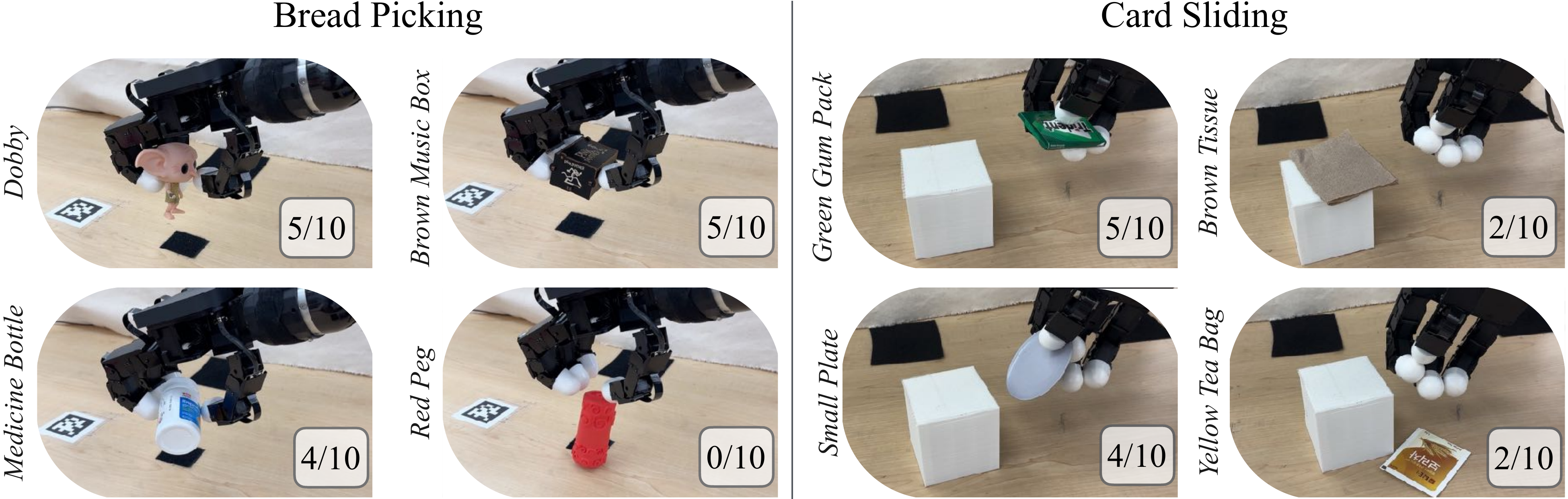}
    \caption{Generalization experiments on the Bread Picking and Card Sliding tasks. We use the text prompts shown on the left side of each image as input to the language-grounded SAM model to generate the initial mask for each object.}
    \label{fig:generalization}
    \vspace{-2mm}
\end{figure*}

Fig. \ref{fig:online_correction} demonstrates how online learning improves the policy in the Paper Sliding task. As can be seen, \method{} enables the robot policy trajectory to move progressively closer to that of the human expert.
To showcase the importance of online corrections, we implement and run the state-of-the-art transformer-based behavior cloning (BC) algorithm VQ-BeT~\cite{lee2024vqbet}, as the base architecture for all of our offline baselines. Similar to the residual policy, the centroid of the tracked points set $\hat{P}^t$, the rotation and translation of the object $\mathcal{T}^t_R$, and the robot's fingertip positions $s^t$ are given as inputs to each BC baseline. We ablate the input and the amount of demonstrations used to experiment on different aspects, and compare \method{} against the following offline baselines:
\begin{enumerate}

    \item \textbf{BC - 1 Demo}~\cite{lee2024vqbet}: \method{} enables training robust policies with only a single human demonstration. For fairness, we compare its offline counterpart and train VQ-BeT end-to-end using only a single demonstration per task. 
    \item \textbf{BC}~\cite{lee2024vqbet}: We run VQ-BeT similar to the previous baseline, but we use 30 demonstrations for each task.    
    \item \textbf{Point Cloud BC}: Similar to DexCap~\cite{wang2024dexcap}, we include point cloud in our input space and modify the input of the BC algorithm by concatenating the point cloud representations received from PointNet~\cite{qi2017pointnet} encoder. We uniformly sample 5000 points from the point cloud and pass them to PointNet without any further preprocessing. Gradients are backpropagated through the entire system, including the point cloud encoder.
    
\end{enumerate}

\begin{table}[ht]
\caption{Comparison of \method{} to different offline algorithms. Paper sliding success is measured in cm of rightward motion; other tasks show success rates out of 10 robot rollouts.}
\centering
\begin{tabular}{@{}ccccc@{}}
\toprule
Method & Bread   & Card & Music & Paper \\ 
       & {\footnotesize (./10)}  & {\footnotesize (./10)} & {\footnotesize (./10)} & {\footnotesize (cm)} \\ \midrule
BC - 1 Demo~\cite{lee2024vqbet}    & 0 & 0 & 0 & 3.5 $\pm$ 1.1 \\  
BC ~\cite{lee2024vqbet}    & 3 & 0 & 0   & 4.1 $\pm$ 1.3  \\
Point Cloud BC~\cite{wang2024dexcap} & 3 & 0 & 0 & 12 $\pm$ 1.3 \\
\method{} (ours)  & \textbf{8} & \textbf{7} & \textbf{6}  & \textbf{17.3 $\pm$ 1.5} \\ \bottomrule
\end{tabular}
\label{tab:policy_results}
\end{table}

Table \ref{tab:policy_results} shows the comparison results. 
As expected, the BC-1 Demo baseline quickly overfits and fails across all tasks. The BC baseline performs relatively well on the Bread Picking task, where precision is less critical. However, on tasks like Card Sliding and Music Box Opening, which require more dexterity, it also overfits quickly—reaching the objects but failing to maintain the necessary consistency. Both BC baselines manage to reach the paper but do little beyond that for Paper Sliding. Our strongest baseline, Point Cloud BC, performs relatively well on Paper Sliding and Bread Picking. We observed that with this baseline, when the robot hand occupies too much of the scene, the data goes out of distribution, causing the model to fail. It moves the paper to the middle but doesn't move it further, grasps the bread but fails to lift it, and reaches for the card but fails to consistently slide it with the thumb. We showcase the failure modes of our most successful offline baseline Point Cloud BC algorithm in Fig. \ref{fig:failure_mode} for three of our tasks. Videos of these failure cases can be seen on our project website.

These results indicate that the dexterous skills learned using this human data collected with \method{} can scale better with more data for some tasks. However, for highly precise tasks such as Music Box Opening and Card Sliding, all offline methods fail, highlighting the importance of online corrections for tasks requiring high dexterity.

\subsection{Does \method{} improve over other reward functions?}
\label{sec:experiments:reward}

In \method{} to compensate for the visual differences between the human and robot videos, we use object-oriented point tracking-based reward functions to guide the learning. We ablate over our design decision, and train online policies for our tasks with the following reward functions:

\begin{enumerate}
    \item \textbf{Image OT}~\cite{haldar2023fish}: We pass RGB images from both the robot and the human videos through pretrained Resnet-18~\cite{he2016deep} image encoders to get image representations and apply optimal transport (OT) based matching on them to get the reward, similar to FISH~\cite{haldar2023fish}. 
    \item \textbf{Point OT}: Instead of direct images we apply OT matching on the points that are tracked throughout both the trajectories of tracked sets of points $\tau_r^p \text{ and } \tau_h^p$.
\end{enumerate}

\begin{figure*}[!t]
    \centering
    \includegraphics[width=\linewidth]{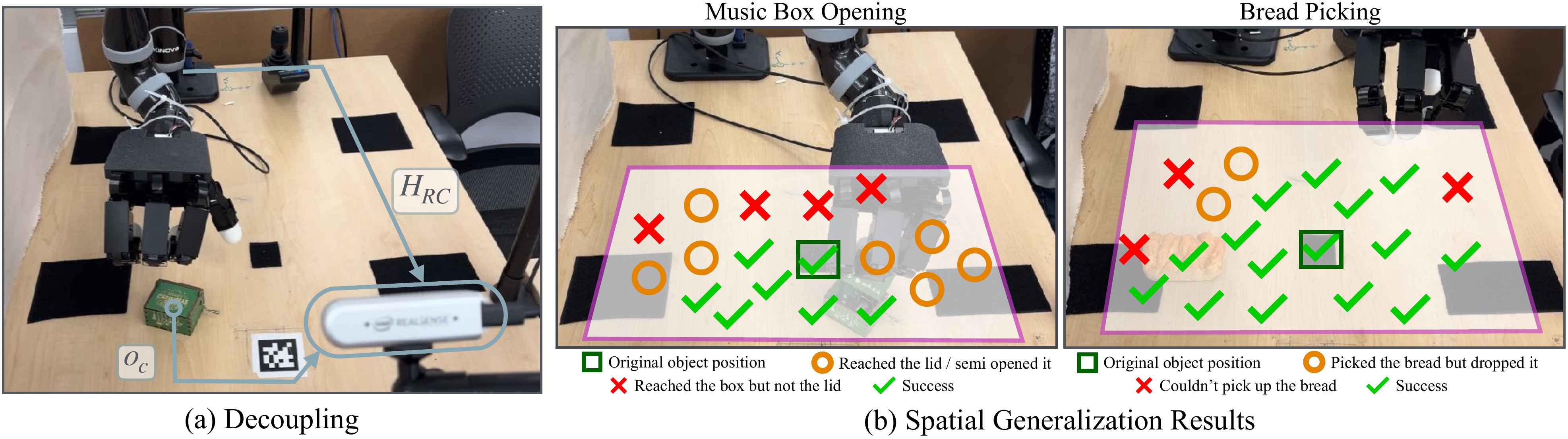}
    \caption{Illustration of how we conduct spatial generalization experiments and their results. First, (a) we decouple the hand and arm generalization by identifying the object location with respect to the robot base and using this location to extract a larger offset to be added to the trained policies with \method{}. Next, (b) we present spatial generalization experiments for the \textit{Bread Picking} and \textit{Music Box Opening} tasks. We illustrate the evaluation locations and report their success below. Note that while policies learned with \method{} generalize successfully in the majority of locations for Bread Picking, they fail to perform similarly in the more dexterous task, Music Box Opening. Videos of evaluation runs can be found on our website.}
    \label{fig:larger_generalization}
\end{figure*}

\begin{table}[ht]
\caption{Comparison of success of \method{} to different reward extraction algorithms. Success is shown as similar to Table~\ref{tab:policy_results}}
\centering
\begin{tabular}{@{}cccc@{}}
\toprule
Method & \method{}(ours)   & Image OT~\cite{haldar2023fish}  & Point OT   \\
\midrule
 Bread Picking  & \textbf{8} & 6  &  6 \\
 Music Box Opening  &  \textbf{6} & 1    & 2   \\
 Paper Sliding (cm) & \textbf{17.25 $\pm$ 1.47} &  16.1 $\pm$ 1.37  & 16.5 $\pm$ 1.23 \\ \bottomrule
\end{tabular}
\label{tab:reward_results}
\end{table}

We present the success rates in Table \ref{tab:reward_results}. We observe that in tasks where the object occupies a large area in the image and the visual differences between the hand and the robot do not significantly affect the image, the Image OT baseline performs similarly to \method{}, as seen in the Paper Sliding task. However, in tasks where the camera needs to be closer to the object to detect its trajectory—such as Music Box Opening—the visual differences between the hand and the robot significantly hinder training, causing image-based reward calculations to fail. On the other hand, direct matching on the predicted points fails because the points tracked for two separate trajectories do not correspond to each other; the same indexed point in one trajectory may correspond to a different location on the object in another trajectory. These differences cause matching to give inconsistent rewards emphasizing the importance of matching \textit{trajectories} rather than points or images. 

\subsection{How well does \method{} generalize to new objects?}
\label{sec:experiments:generalization}

We test the generalization capabilities of policies trained with \method{} on new objects for two tasks, with the results shown in Fig. \ref{fig:generalization}. In these experiments, we apply the policies directly to new objects without retraining, using different text prompts for each inference to obtain object segmentation. We observe that \method{} can generalize with varying success to new objects, provided their shape and texture do not differ significantly from the original. In the Card Sliding task, factors like weight and thickness affect success; for instance, lightweight objects, such as the Yellow Tea Bag, may not fall onto the supporting fingers after sliding, while thin objects, like the Brown Tissue, cannot be slid properly by the thumb. In Bread Picking, \method{} performs well with the Dobby sculpture, despite its different shape from the bread, but fails with slippery objects, like the Red Peg.

These experiments demonstrate that while point-tracking can enable some degree of policy generalization, it is insufficient to overcome substantial differences in shape and texture.

\subsection{How effectively does \method{} generalize to larger areas?}
\label{sec:experiments:larger_generalization}

We evaluated how well policies trained with \method{} generalize spatially to larger areas. Here, we did not train any new policies from scratch for broader spatial generalization due to the extensive online training time required. Instead, we decouple the spatial generalization component from policy learning as follows: we first calculate the object's location in pixel space from the camera using a similar object detection pipeline as described in Section \ref{sec:method:policy_learning}. Then, combined with the depth image, we reproject this to a 3D translation relative to the camera, \( o_c \). We transform this translation to the robot's base frame by \( o_r = H_{RC}^{-1} ~ o_c \) where \( H_{RC} \) is the homogeneous transform from the robot's base to the camera frame as shown in Fig.~\ref{fig:larger_generalization}. Next, we calculate the offset of this location from the original object location in the demonstration $\hat{o}_r$, as \( o_r - \hat{o}_r \). Finally, at inference time, we load the weights for each task trained with \method{} and apply this offset to all the fingers.

\begin{table}[ht]
\caption{The success of spatial generalization. The first column lists the total evaluations for each task, the second describes the behavior, and the third shows successful completions for the given behavior. Note that behaviors are sequential; for example, stabilizing the music box lid can only follow opening it.}
\centering
\begin{tabular}{@{}ccc@{}}
\toprule
\textbf{Task} & \textbf{Behavior} & \textbf{\begin{tabular}[c]{@{}c@{}}Successful \\ Evaluations\end{tabular}} \\ 
\midrule
 & Reached the bread & 20  \\
Bread Picking (./20) & Picked the bread up & 17 \\
 & Held it firmly & 15 \\ 
\midrule 
 & Reached the music box & 18 \\
Music Box Opening & Reached the lid & 14 \\
(./18) & Opened the lid & 10 \\
 & Stabilized the lid & 7\\ 
\bottomrule
\end{tabular}
\label{tab:larger_generalization}
\vspace{-5pt}
\end{table}

We present the results in both Fig. \ref{fig:larger_generalization} and Table \ref{tab:larger_generalization}. While this approach enables fairly effective spatial generalization for the bread-picking task, it fails to generalize for music box opening due to the dexterity and sensitivity required. We observe a trend where the fingers move too high, missing the lid on the leftmost edges of the evaluation area, and too low, lifting the box instead of opening the lid on the rightmost edges. We believe this issue arises both from residual policy predictions and noise in the depth received from the camera.

\vspace{-0.02in}
\section{Limitations and Discussion}

In this paper, we introduced \method{}, a point-tracking, object-oriented reward mechanism designed to close the gap in human-to-robot policy transfer for dexterous hands. \method{} improves upon both offline methods and online counterparts with different reward functions. We also demonstrate \method{}'s generalization to new objects and locations.

Despite its strengths, we identify three limitations. First, our framework only works with in-scene human videos. We believe integrating in-the-wild data collection would significantly enhance its generalization potential. Second, the exploration mechanism requires prior knowledge of which subset of action dimensions is suitable for exploration. Finally, there is no retry mechanism during an episode; when the robot makes a mistake, it can only retry in the next episode. This makes training for long-term tasks challenging. Incorporating a multi-stage learning framework could address this issue. These represent interesting opportunities for future improvements to \method{}.


\section*{Acknowledgements}
We thank Siddhant Haldar, Mara Levy, Jeff Cui, Gaoyue Zhou, Aadhithya Iyer and Venkatesh Pattabiraman for valuable feedback and discussions. This work is supported by grants from Honda, Hyundai, NSF award 2339096 and ONR awards N00014-21-1-2758 and N00014-22-1-2773. LP is supported by the Packard Fellowship.

\bibliographystyle{IEEEtran}
\small
\bibliography{ref}


\end{document}